\pdfoutput=1

\documentclass[11pt]{article}

\usepackage[final]{acl}

\usepackage{times}
\usepackage{latexsym}
\usepackage{amsmath}
\usepackage{setspace}
\usepackage{booktabs}
\usepackage{graphicx} 
\usepackage{hyperref}
\usepackage{latexsym}
\usepackage[T1]{fontenc}
\usepackage[utf8]{inputenc}
\usepackage{microtype}
\usepackage{inconsolata}
\usepackage{multirow}
\usepackage{tabularx}
\usepackage{array}
\usepackage{adjustbox}
\usepackage{multirow}
\usepackage{makecell}
\usepackage{amsfonts}
\usepackage{amssymb}
\usepackage[prependcaption,textsize=tiny]{todonotes}

\usepackage[T1]{fontenc}

\usepackage[utf8]{inputenc}

\usepackage{microtype}
\microtypecontext{spacing=nonfrench}

\usepackage{inconsolata}

\usepackage{graphicx}

\title{Mah\={a}n\={a}ma: A Unique Testbed for Literary Entity Discovery and Linking}

\author{
 \textbf{Sujoy Sarkar\textsuperscript{1}},
 \textbf{Gourav Sarkar\textsuperscript{1}},
 \textbf{Manoj Balaji Jagadeeshan\textsuperscript{1}},
 \textbf{Jivnesh Sandhan\textsuperscript{2}},
\\
 \textbf{Amrith Krishna\textsuperscript{3}},
 \textbf{Pawan Goyal\textsuperscript{1}},
\\
\\
 \textsuperscript{1}Indian Institute of Technology, Kharagpur,
 \textsuperscript{2}Kyoto University, Japan,
 \textsuperscript{3}BharatGen
\\
 \small
 {
 \texttt{sujoys@iitkgp.ac.in, sarkargourav12@gmail.com, manojbalaji1@gmail.com, jivnesh@i.kyoto-u.ac.jp}
 }
 \\
\small
 {
 \texttt{krishnamrith12@gmail.com, pawang.iitk@gmail.com}
 }
}

\begin{document}
\maketitle
\begin{abstract}
High lexical variation, ambiguous references, and long-range dependencies make entity resolution in literary texts particularly challenging. We present \textit{Mahānāma}, the first large-scale dataset for end-to-end Entity Discovery and Linking (EDL) in Sanskrit, a morphologically rich and under-resourced language. Derived from the \textit{Mahābhārata}, the world’s longest epic, the dataset comprises over 109K named entity mentions mapped to 5.5K unique entities, and is aligned with an English knowledge base to support cross-lingual linking. The complex narrative structure of \textit{Mahānāma}, coupled with extensive name variation and ambiguity, poses significant challenges to resolution systems. Our evaluation reveals that current coreference and entity linking models struggle when evaluated on the global context of the test set. These results highlight the limitations of current approaches in resolving entities within such complex discourse. \textit{Mahānāma} thus provides a unique benchmark for advancing entity resolution, especially in literary domains. \footnote{\textit{Mahānāma} is publicly available at \url{https://github.com/sujoysarkarai/mahanama}}

\end{abstract}

\section{Introduction}

The task of Entity Discovery and Linking (EDL) must address two fundamental linguistic challenges: \textit{variability} and \textit{ambiguity} \cite{Tsai2024, Rao2013}. Variability refers to using different expressions to refer to the same entity, while ambiguity arises when the same expression may refer to different entities depending on the context.  Successfully resolving such mentions demands a holistic understanding of discourse within or across documents \cite{zhou-choi-2018-exist}. Most studies on EDL focus on solving these challenges for named entities (NE)~\cite{Tsai2024}. NEs are the central units around which document contents are organised, and accurate resolution is essential for understanding the knowledge expressed in text. Resolving named entities has been shown to enhance representation learning \cite{botha-etal-2020-entity-100-languages}, leading to improved performance in downstream applications such as question answering~\cite{fevry-etal-2020-entities} and knowledge extraction~\cite{chen-etal-2021-evaluating}.

To address the challenges of \textit{variability} and \textit{ambiguity} in EDL, the task is often tackled using end-to-end Entity Linking (EL) systems, which decompose the problem into two sub-components: \textit{mention detection} and \textit{entity disambiguation} \cite{ayoola-etal-2022-refined}. Mention detection identifies spans of text that refer to entities, while entity disambiguation resolves to entries in a knowledge base (KB). A related approach is coreference resolution (CR), which clusters mentions referring to the same entity within a document, without grounding them in a KB \cite{lee-etal-2017-end-to-end}. The two approaches are mutually beneficial \cite{arora-etal-2024-contrastive, bai-etal-2021-joint, durrett-klein-2014-joint}, and a strong cross-document coreference system could, in theory, solve EDL without a KB \cite{Tsai2024}. 

However, entity resolution can be challenging in domains with high lexical variation and contextual ambiguity, particularly in literary corpora \cite{fantasycoref-han-etal-2021-fantasycoref,litbank-bamman-etal-2020-annotated}. Literary texts differ markedly from non-fictional texts like news or Wikipedia: they span long narratives, employ evolving entities and metaphorical expressions, and shift between narrative perspectives \cite{literary-text-roesiger-etal-2018-towards}.  This complexity requires deeper context modeling. Yet, most EDL research remains focused on non-literary domains such as Wikipedia \cite{ghaddar-langlais-2016-wikicoref,botha-etal-2020-entity-100-languages}, news \cite{limkonchotiwat-etal-2023-mrefined}, and web articles \cite{conll-pradhan-etal-2012-conll}, primarily in English, leaving the challenges presented by literary texts and low-resource languages underexplored.

\begin{figure*}[tbh]
    \centering
    \includegraphics[width=\textwidth]{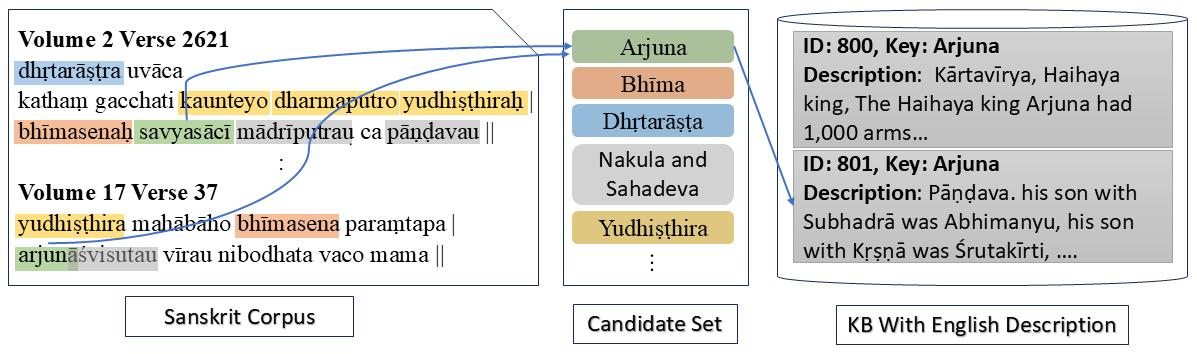}
    \caption{The figure illustrates the structure of our dataset, where name variations are highlighted in the same color. For example, \textit{savyasācī} and \textit{Arjuna} both refer to the same entity, \textit{Arjuna}. Each mention is mapped to an entity, which is linked to an English knowledge base (KB) providing descriptive context. This helps distinguish between different figures sharing the same name, such as two distinct \textit{Arjuna} entries.}
    \label{fig:example}
\end{figure*}

In this work, we present \textit{Mahānāma}\footnote{Derived from \textit{Mahā} (Great) and Nāma (Names), signifying the extensive names in the \textit{Mahābhārata}.}, a dataset constructed from the \textit{Mahābhārata} \cite{krishna1895mahabharata}, the longest epic in world literature, written in Sanskrit, a low-resource and morphologically rich language \cite{sanskrit-graphbased-10.1162/coli_a_00390}. The dataset is derived from a single canonical version of the text and encompasses multiple interwoven narratives, structured in a frame-tale format (stories within a story) \cite{wacks2007framing}. We marked 73K verses using annotation information extracted from the \textit{"Index to the Names in Mahabharata"}\cite{sorensen1904index}, an existing lexicon of names in the epic. The resulting dataset includes 109K mentions spanning 5.5K entities.

Our dataset underscores the core challenges of entity resoltion. NEs in the text display significantly more variability and ambiguity than existing literary datasets. 
For instance, the protagonist \textit{Arjuna} appears under 126 distinct names, while three different characters bear the same name. As shown in Figure \ref{fig:example}, a single verse refers to \textit{Yudhiṣṭhira} using three different names: \textit{kaunteyo}, \textit{dharmaputro}, and \textit{yudhiṣṭhiraḥ}. Some entities, such as \textit{Śiva}, have over one thousand distinct name forms. Such variation is often deeply tied to contextual and cultural cues. Characters are frequently referred to by highly context-dependent names requiring nuanced interpretation. For instance, \textit{Arjuna} is called \textit{Savyasācī} ("ambidextrous") to highlight his unique archery skills, and \textit{Aindri} ("son of Indra") to indicate his divine parentage. These names may not share any lexical similarity, making their resolution especially challenging \cite{moosavi-strube-2017-lexical}. 

Sanskrit also introduces unique linguistic complexities. Words exhibit significant surface-form variation due to inflection and phonetic transformations at boundaries (sandhi), and its verse structure allows relatively free word order \cite{sanskrit-graphbased-10.1162/coli_a_00390, hellwig-nehrdich-2018-sanskrit}. For instance, in Example 1, the span \textit{arjunāśvisutau} refers to three entities: \textit{Arjuna} individually and \textit{Nakula} and \textit{Sahadeva} together. Here, phonetic transformation at the boundary merges \textit{arjuna} and \textit{aśvisutau}, altering \textit{a} into \textit{ā}. 
\begin{flushleft}
\begin{itemize}
    \item[] \textit{arjuna} + \textit{aśvisutau} \(\overset{\textit{a + a = ā}}{\longrightarrow}\) \textit{arjunāśvisutau}
\end{itemize}
\end{flushleft}

Alongside the annotated corpus, we built an English KB with entity descriptions to enable cross-lingual linking between Sanskrit and English. Figure~\ref{fig:example} shows two distinct characters named \textit{Arjuna} from this KB, highlighting the challenge of linking across linguistically distant languages. Multilingual entity linking (MEL) resources remain scarce, with most work focusing on disambiguation rather than end-to-end processing \cite{botha-etal-2020-entity-100-languages}.

Overall, this dataset provides a unique vantage point for analyzing EDL in settings marked by high lexical variability and ambiguity, offering a valuable resource for developing and evaluating more robust resolution systems. The following are the contributions of our work.

\begin{itemize}
\item We present \textit{Mahānāma}, a large literary dataset for Entity Discovery and Linking in Sanskrit, a low-resource and morphologically rich language. The dataset contains 109K annotated mentions over 5.5K entities and captures the core challenges of EDL, namely extreme lexical variation and ambiguity. It is also accompanied by an KB with entity descriptions in English, enabling cross-lingual linking.

\item We compare Mahānāma with existing literary datasets across languages and show that it exhibits substantially higher degrees of lexical and surface-form variation and ambiguity. These characteristics pose significant challenges for current entity resolution systems.

\item We conducted a manual annotation experiment involving annotators with varying familiarity with the \textit{Mahābhārata}. Those with domain-specific background showed higher agreement with our annotations derived from the lexicon than those with only Sanskrit proficiency, suggesting that effective resolution in this dataset requires deep contextual understanding beyond basic linguistic knowledge.

\item We study how variability, ambiguity, and long contextual dependencies in our dataset impact entity resolution by evaluating coreference models, including a mention-ranking baseline \cite{otmazgin-etal-2023-lingmess} and a model designed for long texts \cite{dualcache-guo-etal-2023-dual}. The best F1 of 51.57\% highlights the difficulty of resolving context-dependent names distributed across extended narratives.

\item We also assess an end-to-end multilingual entity linking model \cite{limkonchotiwat-etal-2023-mrefined} that uses entities list, cross-lingual descriptions, and type information. While disambiguation reaches 93.27\% F1 with gold mentions, overall F1 drops to 64.19\% due to mention detection, showing the limits of current models in complex literary settings.

\end{itemize}

\section{Related Work} \label{sec:relatedwork}
The recent rise in interest in literary corpora for entity resolution has underscored challenges such as long documents, narrative complexity, and lexical variation, which are less prominent in standard datasets like AIDA \cite{hoffart-etal-2011-robust} and OntoNotes \cite{conll-pradhan-etal-2012-conll}.

Several corpora have been introduced to address these challenges. The DROC dataset \cite{krug2018droc} contains coreference annotations for 90 German novels with over 393K tokens. LitBank \cite{litbank-bamman-etal-2020-annotated} annotates the first 2,000 tokens of 100 English novels across six entity types. FantasyCoref \cite{fantasycoref-han-etal-2021-fantasycoref} covers 211 fairy tale texts. OpenBoek \cite{openboek} provides 103K tokens corpus from classic Dutch novels, along with spelling normalization to account for historical language variation. KoConovel \cite{kim2024koconovel} focuses on 50 full-length Korean short stories, emphasizing literary resolution in underrepresented languages. Additionally, recent initiatives like CorefUD \cite{nedoluzhko-etal-2022-corefud} have introduced standardized multilingual coreference annotations that includes religious literary text, the Bible. Some datasets focus specifically on named entities. \citet{he-etal-2013-identification} annotate proper names in Pride and Prejudice, while \citet{dutchcoref-van-Zundert} annotate character aliases in 170 Dutch novels, focusing solely on name-based identity resolution and excluding nominals and pronouns. The Friends TV show script corpus \cite{chen-etal-2017-robust}, in contrast, includes over 15K mentions across 46 episodes and supports both CR and EL. But, these datasets do not provide links to any external KBs. 

EL and CR both begin with mention detection, but differ in how they address variation and ambiguity. EL approaches typically handle name variation through alias expansion and candidate generation \cite{Rao2013, ozge-Sevgili}, relying on knowledge base disambiguation supported by entity types, descriptions, and alias lists \cite{ayoola-etal-2022-refined, botha-etal-2020-entity-100-languages}. However, they often struggle with long-tail entities and NIL cases where no matching entry exists \cite{arora-etal-2024-contrastive}. CR models refer to it as lexical variation, encompassing named, nominal, and pronominal mentions, and address it through contextual modeling within the document. Yet their performance declines with increasing document length and lexical diversity \cite{joshi-etal-2019-bert, toshniwal-etal-2020-learning-to-ignore, dutchcoref-van-Zundert, arora-etal-2024-contrastive}. Ambiguity also knows as polysemous mentions, remains a persistent challenge for both tasks \cite{Tsai2024}. Cross-lingual EL is even less explored; Mewsli-9 \cite{botha-etal-2020-entity-100-languages} offers a multilingual benchmark, but is limited to newswire and centers on English as the pivot language.

To address these challenges, we present \textit{Mahānāma}, a novel dataset for evaluating Entity Discovery and Linking in long, complex literary narratives with extensive name variation and ambiguity. It also fills a critical gap as the first large-scale resource for entity resolution in Sanskrit. 

\section{Dataset Creation}
In this section, we present an overview of the resources used for dataset development, detail the manual efforts involved in the creation process, and describe the annotation types.

\subsection{Source}
\textbf{Index:} Our source of annotation is a book, \textit{An Index to the Names in the Mahābhārata}, by Søren Sørensen \cite{sorensen1904index}. This index is a foundational reference for \textit{Mahābhārata} studies, offering a structured catalog of names appearing in the epic. It contains approximately 12.5K primary entries, with many entries listing name variations of entities, expanding the total to around 18K names for entities. The index focuses on proper names, providing verse-level references across the 18 volumes of the \textit{Mahābhārata}. 

We utilized a digitized version of Sørensen's Index\footnote{\url{https://www.sanskrit-lexicon.uni-koeln.de}} \cite{Cologne_Digital_Sanskrit_Dictionaries}. While the resource made the text computationally accessible, it required substantial extraction and manual correction to convert into usable annotation. Sørensen’s Index provides verse references and English descriptions detailing entities and contextual roles within the \textit{Mahābhārata}. We automatically extracted volume and verse numbers from the descriptions and retrieved all name variants linked to each entity. These clusters were then manually reviewed to ensure accurate grouping of name variants. The descriptions were used to construct a cross-lingual knowledge base (KB). Example~\ref{fig:example} shows the descriptions of two entities in the KB.

\textbf{Corpus:} Multiple editions of the \textit{Mahābhārata} exist due to its oral transmission and regional manuscript variations. Sørensen’s Index refers to the Calcutta Edition (CE), which is not digitized and thus cannot be used directly. A digitized OCR version of M.N. Dutta’s 1890s English translation \cite{krishna1895mahabharata}, based on the CE, is available through the \textit{Itihāsa} corpus\footnote{\url{https://github.com/rahular/itihasa}} \cite{aralikatte-etal-2021-itihasa}. However, Dutta’s text introduces structural modifications—merging and splitting verses, rearranging sequences, and inserting or omitting words—causing misalignment with the original. To address this, we undertook a substantial manual effort to align the 73K verses in the digitized text with the 91K verse numbers of the CE. This involved manually reading both editions and assigning CE verse numbers to the corresponding \textit{Itihāsa} verses. Further details are provided in Appendix~\ref{sec:appendix-2}. Table~\ref{tab:mahabharata_structure} shows an overview of the text's structure and structural difference between both editions. 

\begin{table}[t]
\centering
\small
\setlength{\tabcolsep}{3pt}  
\renewcommand{\arraystretch}{0.9}  
\begin{tabular}{lcc}
\toprule
\textbf{Structural Element} & \textbf{CE} & \textbf{M.N. Dutta} \\ 
\midrule
Volumes         & 18  & 9   \\ 
Chapters        & 96  & 157 \\ 
Subchapters     & 2110 & 2110 \\ 
Verses          & 91K & 73K  \\ 
\bottomrule
\end{tabular}
\caption{Structure overview of the \textit{Mahābhārata} (Calcutta Edition and M.N. Dutta)}
\label{tab:mahabharata_structure}
\end{table}

\subsection{Annotation}
\textbf{Entities:} The \textit{Mahābhārata} features a vast array of entities embedded within its narrative. Sørensen’s Index identifies approximately 5.5K unique entities. We manually classify these entities using the CoNLL NER tagset \cite{conll-ner-tjong-kim-sang-de-meulder-2003-introduction} into Person, Location, and Miscellaneous categories  (see Appendix~\ref{sec:appendix-entity-types} for examples). Table \ref{tab:ent_categories} provides distribution of these entity types.

\begin{table}[t]
\centering
\small
\setlength{\tabcolsep}{3pt}  
\renewcommand{\arraystretch}{0.9}  
\begin{tabular}{lcc}
\toprule
\textbf{Category} & \textbf{Entities} & \textbf{Mention \%} \\ 
\midrule
Person       & 4.3K  & 91.1\%  \\ 
Location     & 0.8K  & 3.8\%   \\ 
Miscellaneous & 0.4K  & 5.1\%   \\ 
\bottomrule
\end{tabular}
\caption{Entity distribution across categories}
\label{tab:ent_categories}
\end{table}

\textbf{Mentions:}
A mention is a linguistic expression referring to an entity in discourse \cite{jm3}, including name variations and inflections. In classical Sanskrit literature, distinguishing proper names from nominals is challenging due to frequent use of compounds and derivative phrases as names, often expressing descriptions or relations \cite{sarkar-etal-2023-pre}, making them highly context-dependent. In our dataset, only names identified by the index are annotated as mentions; pronouns (e.g., \ref{fig:example}, \textit{mama} “my”) and common nouns (e.g., \ref{fig:example}, \textit{vīrau} “two warriors”) are excluded.

The corpus is unsegmented and contains multi-word tokens (MWTs) \cite{nivre-etal-2017-universal}, where multiple words are joined together through phonological merging (sandhi) and compounding \cite{sanskrit-graphbased-10.1162/coli_a_00390}. These MWTs often include more than one entity mention, with 39\% of mentions in our dataset occurring within such merged forms. We annotate mention boundaries within each verse at the character level. To assist in segmenting these MWTs and identifying the start and end of inflected names, we use two tools: the Sanskrit Heritage Reader \cite{shr-Goyal_Huet_2016}, a lexicon-based shallow parser, and a neural network–based segmenter \cite{hellwig-nehrdich-2018-sanskrit}. For a detailed explanation of this process, please refer to Appendix \ref{sec:appendix-2}. For example, in the MWT \textit{arjunāśvisutau}, two mentions are embedded: \textit{arjuna}\textsubscript{1} and \textit{aśvisutau}\textsubscript{2}, which we annotate as:

\begin{flushleft}
\begin{itemize}
\item[] \textit{arjunāśvisutau} $\xrightarrow{\text{Boundary}}$ \textit{arjunā}\textsubscript{1}, \textit{āśvisutau}\textsubscript{2}
\end{itemize}
\end{flushleft}

\textbf{Clusters and Knowledge Base:}
Two or more mentions referring to the same entity within a discourse are considered coreferential \cite{jm3}. All occurrences of an entity name, including its name variations, are grouped into a single cluster, identified by a unique cluster ID. In addition, each cluster is linked to the KB, which provides cross-lingual descriptions in English.

\textbf{Special Considerations:}
Our dataset explicitly marks appositive and copular mentions within the same coreference cluster, following approaches from Preco and KocoNovel \cite{preco-chen-etal-2018-preco, kim2024koconovel}. Dual and plural mentions are linked only to mentions of the same grammatical number, as per OntoNotes guidelines \cite{agarwal-etal-2022-entity-linking}. Nested entities within proper names are not annotated separately to maintain consistency with prior work \cite{kim2024koconovel}. We also include singleton entities, aligning with LitBank and Preco \cite{litbank-bamman-etal-2020-annotated, preco-chen-etal-2018-preco}, ensuring comprehensive entity coverage. Further details on these are provided in Appendix \ref{sec:appendix-1}.

\subsection{Inter-Annotator Agreement}

\begin{table}[ht]
\small
\setlength{\tabcolsep}{3.5pt} 
\centering
\begin{tabular}{|c|c|c|c|c|}
\hline
\multicolumn{2}{|l|}{\textbf{\textit{Mahānāma} Annotation vs.}} & \makecell[c]{Expert\\1} & \makecell[c]{Expert\\2} & \makecell[c]{Non-expert\\(Avg)} \\
\hline 
\multirow{2}{*}{Span} 
  & $\kappa$      & 0.91 & 0.86 & 0.76 \\
  & F1            & 0.92 & 0.87 & 0.78 \\

\hline
\rule{0pt}{2.0ex}
\multirow{3}{*}{\makecell{Span\\+\\Link} } 
  & $\kappa$ (All tokens) & 0.89 & 0.81 & 0.69 \\
  & $\kappa$ (Entity tokens)& 0.80 & 0.67 & 0.53 \\
  & F1               & 0.80 & 0.68 & 0.56 \\

\hline
\end{tabular}
\caption{IAA of \textit{Mahānāma} Annotation vs. Expert and Non-expert Annotators; $\kappa$ = Cohen's Kappa}
\label{tab:inter-annotator}
\end{table}

To assess annotation quality and dataset difficulty, we conducted an inter-annotator agreement study on 1,000 randomly sampled verses. Table~\ref{tab:inter-annotator} presents results for both mention detection (Span) and entity linking (Span+Link), comparing our annotations with two Sanskrit experts (both with master’s degrees and expert 1 with prior experience in \textit{Mahābhārata} studies) and a non-expert group (two students with school-level Sanskrit proficiency). We report token-level Cohen’s $\kappa$ for all tokens and entity tokens, and F1 scores excluding non entity tokens, as recommended by \citet{Deleger2012}.

Mention detection showed high agreement across all annotator groups, with Cohen’s $\kappa$ indicating nearly perfect alignment with both experts ($\kappa$ = 0.92, 0.86). When entity disambiguation is included, the task becomes more challenging, as reflected in a wider F1 difference between experts (0.12 for Span+Link vs. 0.05 for Span). Despite this, our annotation achieves a close to near-perfect $\kappa$ of 0.80 with Expert 1 for entity linking, affirming the reliability and domain-informed accuracy of our annotations. These findings suggest that effective resolution in this dataset requires deep contextual understanding beyond basic linguistic knowledge. See Appendix~\ref{sec:appendix-iaa} for details.

\section{Dataset Analysis}
This section presents our dataset's basic statistics, highlighting its unique properties by  comparing it with relevant literary and non-literary entity resolution corpora (introduced in Section~\ref{sec:relatedwork}).

\textbf{Basic Statistics:} Our dataset contains 988,502 white space separeted tokens, making it significantly larger than other public literaray datasets for entity resolution as shown in Table \ref{tab:compact_dataset_overview}. Additionally, our dataset is rich in NEs. Literary corpora typically have higher proportions of pronouns compared to non-literary domains\cite{pagel-reiter-2020-gerdracor}. In our dataset, despite only NEs are marked, 10.56\% of the tokens are identified as mentions, highlighting a notable entity density. 

\textbf{Major Entities}: In literary texts, a few key entities dominate the narrative, making up most mentions \cite{litbank-bamman-etal-2020-annotated,dualcache-guo-etal-2023-dual}. As shown in Table \ref{tab:properties_comaprision}, literary corpora typically have fewer entities than non-literary ones, with under 10\% of entities contributing to over 50\% of mentions. This concentration shapes the primary narrative. In our dataset, we analyze major entities at subchapter, chapter, and corpus levels. When considering the dataset as a whole, only 26 entities account for 50\% of the total mentions.

\begin{table}[h!]
\centering
\small
\renewcommand{\arraystretch}{1.1}
\setlength{\tabcolsep}{5pt}
\resizebox{\columnwidth}{!}{
\begin{tabular}{lcccc}
\toprule
\textbf{Dataset} & \textbf{Docs} & \textbf{Tokens} & \textbf{Mentions} & \textbf{Entities} \\[0.5ex]
\midrule
DROC (Lit.) & 90 & 393K & 52K & 5.3K \\
Litbank (Lit.) & 100 & 210K & 29K & 7.9K \\
Fantasycoref (Lit.) & 214 & 367K & 62K & 6.2K \\
KocoNovel (Lit.) & 50 & 178K & 19K & 1.4K \\
Openboek (Lit.) & 9 & 103K & 23.6K & 8.9K \\
\midrule
OntoNotes (Non-Lit.) & 3493 & 1631K & 194K & 44K \\
Mewsli-9 (Non-Lit.) & 58K & 20M & 289K & 82K \\
\midrule
\textit{Mahānāma} (Lit.) & - & 988K & 109K (Only NE) & 5.5K \\
\bottomrule
\end{tabular}}
\caption{Comparison of basic statistics across literary (Lit.) and non-literary (Non-Lit.) corpora.}
\label{tab:compact_dataset_overview}
\end{table}


\textbf{Lexical Variations:} Our dataset shows significantly higher lexical variation in names of major entities, with an average of 8.69 unique forms per entity at the chapter level and 124.42 at the dataset level (Table~\ref{tab:properties_comaprision}). For comparison datasets, we excluded only pronominal mentions and included both named and nominal forms when computing variation. Even when considering only NEs, our dataset exhibits nearly twice the variation seen in LitBank at the chapter level. At the dataset level, major entity clusters show extreme diversity, with one entity (\textit{śiva}) appearing in up to 1,385 distinct forms. Additionally, our dataset displays exceptionally high surface-form variation due to the nature of the language.

\begin{table*}[ht]
\centering

\begin{adjustbox}{max width=\textwidth}
\begin{tabular}{|l|l|l|c|cc|cc|c|}
\hline
\multirow{3}{*}{\textbf{Dataset Name}} & \multirow{3}{*}{\textbf{Language}} & \multirow{3}{*}{\textbf{Texts}} & \multicolumn{5}{c|}{\textbf{Major Entities (covering 50\% of mentions)}} & \multirow{3}{*}{\makecell{\textbf{Avg. \% entities} \\ \textbf{with ambiguous} \\ \textbf{mentions}}} \\ \cline{4-8} 
                                       &                                    &                                         & \multirow{2}{*}{\makecell{\textbf{\% of} \\ \textbf{total} \\ \textbf{entities}}} & \multicolumn{2}{c|}{\textbf{Lexical Variation (Stem)}} & \multicolumn{2}{c|}{\textbf{Surface Form}} &                                                                   \\ \cline{5-8}
                                       &                                    &                                         &                                         & \multicolumn{1}{c|}{\makecell{\textbf{Avg. \# of} \\ \textbf{variation}}} & \makecell{\textbf{Max. \# of} \\ \textbf{variation}} & \multicolumn{1}{c|}{\makecell{\textbf{Avg. \# of} \\ \textbf{variation}}} & \makecell{\textbf{Max \# of} \\ \textbf{variation}} &                                                                   \\ \hline
DROC                                & German                            & Literary                                & 4.99\%                                    & \multicolumn{1}{c|}{6.63}                        & 29                         & \multicolumn{1}{c|}{\textbf{8.26}}                        & 37                         &  \textbf{36.23}\%                                                              \\ \hline
Litbank                                & English                            & Literary                                & 5.83\%                                    & \multicolumn{1}{c|}{4.02}                        & 20                         & \multicolumn{1}{c|}{4.19}                        & 23                         & 10.0\%                                                              \\ \hline
Fantasycoref                           & English                            & Literary                                & 10.02\%                                   & \multicolumn{1}{c|}{\textbf{6.86}}                           & 33                          & \multicolumn{1}{c|}{7.53}                        & 34                         & 16.0\%                                                              \\ \hline
Openboek                               & Dutch                              & Literary                                & 3.75\%                                    & \multicolumn{1}{c|}{5.26}                           & \textbf{53}                          & \multicolumn{1}{c|}{5.50}                        & 55                         & 25.0\%                                                              \\ \hline
KocoNovel                              & Korean                             & Literary                                & 18\%                                      & \multicolumn{1}{c|}{-}                           & -                          & \multicolumn{1}{c|}{2.4}                         & 14                         & 12.0\%                                                              \\ \hline
CorefUD Proiel                                 & Ancient Greek                      & Bible                                   & 9.50\%                                    & \multicolumn{1}{c|}{5.75}                        & 34                         & \multicolumn{1}{c|}{6.31}                        & 35                         & 27.0\%                                                              \\ \hline
CorefUD Proiel                                 & Old Slavonic                & Bible                                   & 10.70\%                                   & \multicolumn{1}{c|}{4.85}                        & 27                         & \multicolumn{1}{c|}{5.83}                        & 32                         & 28.0\%                                                              \\ \hline \hline
Ontonotes                              & English                            & News, Web                 & 24.69\%                                   & \multicolumn{1}{c|}{-}                           & -                          & \multicolumn{1}{c|}{2.65}                        & 27                         & 2.0\%                                                               \\ \hline \hline
Mewsli-9                              & 11 Languages                            & Wikinews                 & 4.52\%                                   & \multicolumn{1}{c|}{-}                           & -                          & \multicolumn{1}{c|}{5.33}                        & \textbf{57}                         & 11.74\%                                                               \\ \hline \hline
\textit{Mahānāma} (Subch.)                  & Sanskrit                           & Literary                                & 27.56\%                                   & \multicolumn{1}{c|}{2.66}                        & 751                        & \multicolumn{1}{c|}{4.9}                         & 752                        & 6.0\%                                                               \\ \hline
\textit{Mahānāma} (Ch.)                     & Sanskrit                           & Literary                                & 5.17\%                                    & \multicolumn{1}{c|}{8.69}                        & 1021                       & \multicolumn{1}{c|}{27.17}                       & 1078                       & 17.0\%                                                              \\ \hline
\textit{Mahānāma} (Total)                       & Sanskrit                           & Literary                                & 0.46\%                                    & \multicolumn{1}{c|}{124.42}                      & 1385                       & \multicolumn{1}{c|}{640.58}                      & 2187                       & 47.0\%                                                              \\ \hline
\end{tabular}
\end{adjustbox}

\caption{Comparison of dataset properties. Our dataset is analyzed at three levels—Subch (subchapter), Ch (chapter), and Total (entire dataset). For other datasets, variation includes both NE and nominal mentions, while ours is NE-only. "-" indicates low surface-form variation or unavailable stems, so lexical variation was not computed.}

\label{tab:properties_comaprision}
\end{table*}

\textbf{Ambiguity:} Ambiguity poses a major challenge in our dataset. As shown in Table \ref{tab:properties_comaprision}, ancient literary texts such as the Bible exhibit higher ambiguity than non-literary. Notably in our datset  47\% of entities share a common name, requiring context-based disambiguation essential. For example, \textit{Janamejaya} refers to ten distinct characters. The challenge is intensified in Sanskrit, where the lack of clear markers makes it hard to distinguish proper names from common nouns \cite{kim2024koconovel}. As seen in Figure \ref{fig:example}, \textit{mahābāho}(the mighty-armed) is used an adjective for \textit{Yudhishthira}, while \textit{mahābāhu} is also name of other distinct characters.

\begin{figure}[h!]
    \centering
    \includegraphics[width=\columnwidth]{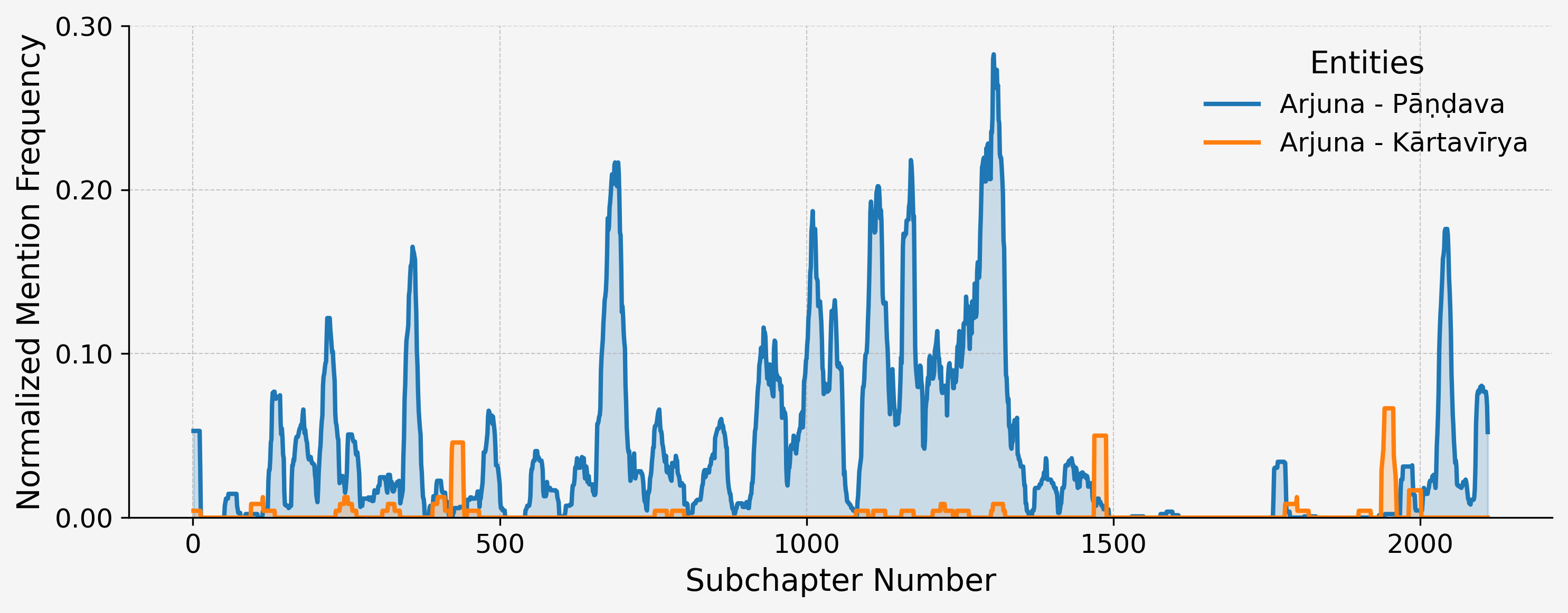}
    \caption{Mention frequency of Arjuna (Pāṇḍava) and Arjuna (Kārtavīrya) across 2K subchapters, illustrating bursty distribution and overlapping spans.}
    \label{fig:entity_distribution}
\end{figure}

\textbf{Spread and Burstiness:} In literary texts, entities often follow a bursty pattern—long spans with few mentions punctuated by periods of intense focus \cite{litbank-bamman-etal-2020-annotated}. Figure \ref{fig:entity_distribution} shows the distribution of \textit{Arjuna} across 2K subchapters, with high-frequency peaks and intermittent gaps. It also highlights a minor, overlapping entity with the same name. Resolution models must handle such burstiness and overlapping spans to accurately link mentions.

\section{Experiments}
We evaluate both coreference resolution (CR) and entity linking (EL) models for the task. In CR, given a document \( D \), the goal is to cluster mentions \( M = \{m_1, \dots, m_{|M|}\} \) into entity clusters \( C = \{c_1, \dots, c_{|C|}\} \) via a function \( f_{\text{CR}}: M \to C \). In EL, with a knowledge base \( KB \) of entities \( E = \{e_1, \dots, e_{|E|}\} \), the task maps mentions to entities using \( f_{\text{EL}}: M \to E \). EL models rely on candidate sets and entity descriptions. We analyze the role of external knowledge and how our dataset enables studying local vs.\ global context in long-form narratives.

\subsection{Models}

As baselines, we evaluate \textbf{LingMess}~\cite{otmazgin-etal-2023-lingmess}, a CR model extending the mention-ranking (MR) architecture of \citet{lee-etal-2017-end-to-end}, which allows us to excludes pronoun-related coreference scorers, making it suitable for our dataset. We also use \textbf{Dual Cache}~\cite{dualcache-guo-etal-2023-dual}, an entity-ranking (ER) model designed for long literary texts, which incrementally processes documents using to capture local and global entities, ideal for our dataset's structure. For multilingual entity linking, we assess \textbf{mReFiNeD}~\cite{limkonchotiwat-etal-2023-mrefined}, a state-of-the-art bi-encoder model leveraging entity types and cross-lingual descriptions, ensuring robust zero-shot capabilities within an academic computational budget.

\subsection{Experiment Settings}
\textbf{Setup:} For LingMess \cite{otmazgin-etal-2023-lingmess}, we disable pronoun-related scorers due to the absence of pronoun annotations. Dual Cache \cite{dualcache-guo-etal-2023-dual} is configured to prevent cache misses with appropriate local and global cache sizes. Both models use Longformer-Large \cite{beltagy2020longformer}. mReFiNeD is trained in a multi-task setting using MuRIL \cite{khanuja2021murilmultilingualrepresentationsindian} for encoding. See Appendix~\ref{sec:appendix-3} for more details.

\textbf{Metric: } For coreference resolution, we use the standard CoNLL scorer, which reports F1 scores for \textbf{MUC}, \textbf{\(\text{B}^3\)}, and \textbf{CEAF\textsubscript{\(\phi_4\)}} \cite{moosavi-strube-2016-coreference-metric}. The final score is the \textbf{average F1} across these metrics. For end-to-end entity linking, we report \textbf{InKB micro-F1} with strict mention boundary matching, requiring exact matches to gold mentions. Mention detection is evaluated separately using \textbf{F1} score.

\textbf{Dataset Division:}
EL and CR models are typically trained at the document level, each representing a single discourse. In our dataset, the entire corpus is treated as one discourse, structured as shown in Table~\ref{tab:compact_dataset_overview}. Each subchapter, averaging 468 tokens, forms a coherent part of the Mahābhārata and serves as an independent training document. The dataset is split into 1,688 subchapters for training, 211 for development, and 211 for testing. Evaluation considers both per-subchapter performance (local) and overall test set performance (global) as a single discourse. The manually annotated 1,000 verses sampled across the text were not used for evaluation, as their scattered nature lacks the narrative context needed.

\textbf{Handling Unsegmented Data:} Most CR models, including the two used in our study are not designed to operate directly on unsegmented text. To address this, we adapt the Dual-Cache models to predict entity boundaries at the subtoken level as it performed better at token level. It enabled better handling of Multi-Word Tokens. This involved modifying the model code to support subtoken-level boundary prediction. For entity linking, we use character-level spans, while for coreference, entity boundaries are derived from tokenizer-generated subtokens. We evaluate both token- and subtoken-level setups to quantify their impact.

\begin{table*}[htbp]
\centering
\small  
\setlength{\tabcolsep}{3pt}  
\begin{tabular}{l*{13}{r}}
\toprule
\makecell{Model} & \makecell{Type} & \makecell{Entity \\ Boundary \\ Marking} & \makecell{Eval. \\ Level} & \multicolumn{3}{c}{MUC} & \multicolumn{3}{c}{$\text{B}^3$} & \multicolumn{3}{c}{CEAF\textsubscript{$\phi_4$}} & Avg. \\
\cmidrule(lr){5-7} \cmidrule(lr){8-10} \cmidrule(lr){11-13}
& & & & P & R & F1 & P & R & F1 & P & R & F1 & F1 \\
\midrule
Lingmess & MR & Token & Local & 
82.30 & 75.90 & \textbf{79.00} & 76.30 & 67.90 & 71.90 & 74.00 & 29.10 & 41.80 & 64.20 \\
Dual-Cache & ER & Token & Local & 
65.52 & 81.31 & 72.57 & 67.05 & 78.67 & 72.40 & 70.54 & 61.35 & 65.63 & \textbf{70.30} \\
\midrule
{\footnotesize Dual-Cache} & ER & Subtoken & Local & 
72.78 & 83.95 & 77.96 & 70.61 & 80.02 & \textbf{75.02} & 75.59 & 67.47 & \textbf{71.30} & \textbf{74.76} \\
{\footnotesize Dual-Cache} & ER & Subtoken & Global & 
67.30 & 84.50 & 74.92 & 37.31 & 67.72 & 48.11 & 48.83 & 23.45 & 31.68 & 51.57 \\
\bottomrule
\end{tabular}
\caption{Performance of the CR models on the test set. Model types: MR = Mention Ranking, ER = Entity Ranking}
\label{tab:results}
\end{table*}

\begin{table}[ht]
    \centering
    \renewcommand{\arraystretch}{1.2} 
    \setlength{\tabcolsep}{5pt} 
    \footnotesize 
    \resizebox{\columnwidth}{!}{%
    \begin{tabular}{l|l|c|c|c}
        \toprule
        \textbf{\raggedright Task} & \textbf{Model} & \textbf{P} & \textbf{R} & \textbf{F1} \\
        \hline
        \multirow{3}{*}{\makecell[l]{Entity\\Linking}} 
        & mReFiNeD & 80.51 & 53.38 & \textbf{64.19} \\
        & w/o descriptions & 79.41 & 52.18 & 62.98 \\
        & w/o entity types & 80.47 & 53.33 & 64.15 \\
        \hline
        \multirow{3}{*}{\makecell[l]{Entity\\Disambiguation}} 
        & mReFiNeD & 93.30 & 93.24 & \textbf{93.27} \\
        & w/o descriptions & 91.55 & 91.25 & 91.40 \\
        & w/o entity types & 93.01 & 93.12 & 93.06 \\
        \hline
        \multirow{2}{*}{\makecell[l]{Mention\\Detection}} 
        & mReFiNeD & 63.06 & 57.63 & 60.22 \\
        \cline{2-5}
        & Dual-Cache & 86.36 & 81.50 & \textbf{83.86} \\
        \bottomrule
    \end{tabular}}
    \caption{Performance of models on Entity Linking, Entity Disambiguation, and Mention Detection.}
    \label{tab:combined_results}
\end{table}

\section{Results}
\subsection{Performance of Coreference Models}

Table~\ref{tab:results} shows CR model results, evaluated both locally (within subchapters) and globally (across the full test set) using token- and subtoken-level mention boundaries. At the token level, DualCache outperforms LingMess with an average F1 of 70.31. LingMess excels on the MUC metric (F1 79.00), which emphasizes linkage accuracy, suggesting better handling of name variations. However, it struggles with entity alignment, as seen in its low CEAF$\phi_4$ F1 (41.80). In contrast, DualCache performs more consistently across metrics. With subtoken-level boundary training, DualCache improves its average F1 by 4.16 points (74.46) and achieves its highest B$^3$ F1 (75.02), showing better mention detection and MWT handling. Globally, DualCache's CEAF$\phi_4$ F1 drops to 31.68, reducing its average F1 to 51.57\%. While MUC remains stable, the CEAF$\phi_4$ drop suggests difficulty in resolving ambiguous entities across the full discourse, highlighting the need for better global resolution.

\subsection{Performance of Entity Linking Model}
Table~\ref{tab:combined_results} presents results for Entity Linking (EL), Disambiguation, and Mention Detection. mReFiNeD, applied globally, achieves an EL F1 of 64.19\%, indicating potentially stronger global performance than CR models, though the scores are not directly comparable. However, its performance is limited by weak mention detection, with an F1 of 60.22\%, significantly lower than DualCache (F1: 83.86\%), highlighting the need to improve end-to-end models. 

Ablation studies show that both cross-lingual descriptions and entity types contribute modestly to EL. Removing descriptions lowers F1 by 1.21 points, while removing types has negligible impact. This suggests that descriptions offer limited contextual benefit for resolving ambiguous entities. For entity disambiguation, which involves resolving ambiguous mentions given gold spans, mReFiNeD performs strongly with an F1 of 93.27 but relies on external resources such as a restricted set of candidates and their prior probabilities, underscoring the need for more self-sufficient approaches. As with EL, ablations show complementary contributions from descriptions and entity types.

\section{Error Analysis}

\begin{table}[t]
\centering
\footnotesize
\renewcommand{\arraystretch}{1.1}
\setlength{\tabcolsep}{3pt}
\resizebox{\columnwidth}{!}{
\begin{tabular}{l c c c c}
\toprule
\textbf{Metric} & \makecell{\textbf{Lingmess}\\\textbf{(Local)}} & \makecell{\textbf{Dual-Cache}\\\textbf{(Local)}} & \makecell{\textbf{Dual-Cache}\\\textbf{(Global)}} & \makecell{\textbf{mReFiNeD}\\\textbf{(Global)}} \\
\midrule
Conf. Ent. \%     & 10.4 & 3.6   & 7.8   & 2.00   \\
Div. Ent. \%      & 11.5 & 10.0  & 33.2  & 5.07   \\
Miss. Ent. \%     & 15.3 & 17.3  & 26.9  & 32.76  \\
Miss. Ment. \%    & 9.1  & 8.9   & 4.7   & 17.6  \\
\hline
Extra Ent. \%     & 20.0 & 15.2  & 37.7  & 16.5   \\
Extra Ment. \%    & 10.4 & 7.2   & 6.0   & 29.2   \\
\bottomrule
\end{tabular}
}
\caption{Automatically identified errors percentage in predictions. \textbf{Conflated Entities:} distinct entities merged; \textbf{Divided Entity:} a single entity split into multiple; \textbf{Missing/Extra Mention/Entity:} mention/entity missing or incorrectly added. Span errors were not considered, as all spans are within single-token.}
\label{tab:error}
\end{table}

\textbf{Qualitative Analysis: }
Both CR and EL models struggle with entity mentions in the \textit{Mahābhārata}. The best-performing CR model fails to link lexical variations, as seen in Volume~1, Chapter~12, Subchapter~190, where the entity \textit{draupadī} appears nine times but is split into three clusters: [\textit{yājñasenī}, \textit{kṛṣṇāṃ}, \textit{yājñasenī}, \textit{yājñasenī}]; [\textit{pāñcālyāṃ}, \textit{pāñcālyā}]; and [\textit{kṛṣṇāṃ}, \textit{draupadī}, \textit{draupadī}], showing a tendency to group mentions based on surface similarity. It also fails to disambiguate ambiguous mentions. In Volume~7, Chapter~6, Subchapter~165, \textit{Bhūri} (son of \textit{Somadatta}) and \textit{Duryodhana} (eldest son of \textit{Dhṛtarāṣṭra}) are both referred to as \textit{kaurava}, yet the model clusters all occurrences under a single entity.

The EL model correctly links all mentions of \textit{draupadī} but struggles with general references. In the same document, it mistakenly links \textit{pārtho} (plural, referring to the sons of \textit{Pṛthā}) to \textit{bhīma} (one of them). Similarly, in another document, \textit{kauravaḥ} is wrongly linked to \textit{duryodhana} instead of \textit{bhūri}, likely due to prior probability bias. The model also struggles with mention boundary detection, especially for MWTs. These issues highlight the need for improved handling of name variations, ambiguity, context-aware resolution, and morphological richness in both approaches.

\textbf{Quantitative Analysis: }
To assess model performance differences, we also conduct an error analysis based on the Berkeley Coreference Analyzer’s error types \cite{kummerfeld-klein-2013-error}, which categorizes errors into seven types. Table \ref{tab:error} presents the error distribution across models, with lower error percentage reflecting stronger performance. Refer to the Appendix \ref{sec:appendix-error} for more details.

\section{Conclusion}
We introduced \textit{Mahānāma}, a large-scale Sanskrit dataset for Entity Discovery and Linking that captures challenges in literary texts, including extreme name variation, contextual ambiguity, and long-range dependencies. Derived from the \textit{Mahābhārata}, the world’s longest epic, it contains 109K mentions across 5.5K entities, annotated using a name index and linked to an English knowledge base. Evaluation of coreference and entity linking models reveals difficulty in resolving name variation and ambiguous mentions over long contexts. \textit{Mahānāma} provides a valuable benchmark for advancing robust, context-aware entity resolution in complex literary settings.

\section*{Limitations}
While \textit{Mahānāma} makes a substantial contribution to Sanskrit entity resolution, certain limitations arise from the nature of its source material and annotation methodology. The annotations were derived automatically from a name index authored by a domain expert, which provides verse-level references without pinpointing exact name occurrences, necessitating a string-matching approach. To ensure high precision, only uniquely identifiable mentions were annotated, potentially omitting some instances. The dataset also inherits some OCR errors from the source corpora, for which no manual correction was attempted. Furthermore, the annotation focuses exclusively on named entities, excluding pronouns and common noun mentions, and is therefore not intended for comprehensive coreference resolution, though it lays the groundwork for future extensions in that direction. The definition of a “name” follows the expert author’s perspective, as no standardized named entity guidelines exist for Sanskrit. While coreferential links were assigned following certain guidelines such as linking dual and plural mentions only to corresponding dual and plural entity forms. Additionally, because the dataset is based on a classical epic presented in verse format, its applicability to modern or prose texts may be limited and would require further investigation using techniques such as poetry-to-prose conversion. Since the training and test sets are drawn from the same narrative, some overlap in main entities is unavoidable, which may result in overestimation of model performance. Nonetheless, the dataset provides a valuable foundation, and future work can build upon it by exploring techniques such as data augmentation.

\section*{Ethics Statement}
The annotations in this work are derived from published, copyright-free sources and a publicly available corpus. All resources utilized have been appropriately cited. The dataset, including annotations, is constructed from existing literary sources, and no explicit bias analysis has been performed. The dataset, annotations and codes is released under a CC-0 license. Annotation mapping was primarily carried out using automated methods, with experts validation conducted to ensure quality assessment and corpus alignment. Manual corpus alignment was performed by two graduate student contributors who studied Sanskrit in school, while a randomly selected set of 1000 verses was annotated by same two students and two expert with a master's degree in Sanskrit and one with a background in \textit{Mahābhārata} studies. Annotators involved in the process were fairly compensated in accordance with standard institutional guidelines. The dataset does not contain any personal or sensitive information.

\section*{Acknowledgements}
We appreciate and thank all the anonymous reviewers for their constructive feedback towards improving this work. The work was supported in part by the National Language Translation Mission (NLTM): Bhashini project by the Government of India.

\section*{AI Assistance}
AI assistants such as Grammarly and ChatGPT were used in the writing process to refine textual clarity and structure.

\bibliography{refs}

\begin{thebibliography}{55}
\expandafter\ifx\csname natexlab\endcsname\relax\def\natexlab#1{#1}\fi

\bibitem[{Agarwal et~al.(2022)Agarwal, Angell, Monath, and McCallum}]{agarwal-etal-2022-entity-linking}
Dhruv Agarwal, Rico Angell, Nicholas Monath, and Andrew McCallum. 2022.
\newblock \href {https://doi.org/10.18653/v1/2022.naacl-main.343} {Entity linking via explicit mention-mention coreference modeling}.
\newblock In \emph{Proceedings of the 2022 Conference of the North American Chapter of the Association for Computational Linguistics: Human Language Technologies}, pages 4644--4658, Seattle, United States. Association for Computational Linguistics.

\bibitem[{Aralikatte et~al.(2021)Aralikatte, de~Lhoneux, Kunchukuttan, and S{\o}gaard}]{aralikatte-etal-2021-itihasa}
Rahul Aralikatte, Miryam de~Lhoneux, Anoop Kunchukuttan, and Anders S{\o}gaard. 2021.
\newblock \href {https://doi.org/10.18653/v1/2021.wat-1.22} {Itihasa: A large-scale corpus for {S}anskrit to {E}nglish translation}.
\newblock In \emph{Proceedings of the 8th Workshop on Asian Translation (WAT2021)}, pages 191--197, Online. Association for Computational Linguistics.

\bibitem[{Arora et~al.(2024)Arora, Silcock, Dell, and Heldring}]{arora-etal-2024-contrastive}
Abhishek Arora, Emily Silcock, Melissa Dell, and Leander Heldring. 2024.
\newblock \href {https://doi.org/10.18653/v1/2024.emnlp-main.355} {Contrastive entity coreference and disambiguation for historical texts}.
\newblock In \emph{Proceedings of the 2024 Conference on Empirical Methods in Natural Language Processing}, pages 6174--6186, Miami, Florida, USA. Association for Computational Linguistics.

\bibitem[{Ayoola et~al.(2022)Ayoola, Tyagi, Fisher, Christodoulopoulos, and Pierleoni}]{ayoola-etal-2022-refined}
Tom Ayoola, Shubhi Tyagi, Joseph Fisher, Christos Christodoulopoulos, and Andrea Pierleoni. 2022.
\newblock \href {https://doi.org/10.18653/v1/2022.naacl-industry.24} {{R}e{F}in{ED}: An efficient zero-shot-capable approach to end-to-end entity linking}.
\newblock In \emph{Proceedings of the 2022 Conference of the North American Chapter of the Association for Computational Linguistics: Human Language Technologies: Industry Track}, pages 209--220, Hybrid: Seattle, Washington + Online. Association for Computational Linguistics.

\bibitem[{Bai et~al.(2021)Bai, Zhang, Song, and Xu}]{bai-etal-2021-joint}
Jiaxin Bai, Hongming Zhang, Yangqiu Song, and Kun Xu. 2021.
\newblock \href {https://doi.org/10.18653/v1/2021.eacl-main.43} {Joint coreference resolution and character linking for multiparty conversation}.
\newblock In \emph{Proceedings of the 16th Conference of the European Chapter of the Association for Computational Linguistics: Main Volume}, pages 539--548, Online. Association for Computational Linguistics.

\bibitem[{Bamman et~al.(2020)Bamman, Lewke, and Mansoor}]{litbank-bamman-etal-2020-annotated}
David Bamman, Olivia Lewke, and Anya Mansoor. 2020.
\newblock \href {https://aclanthology.org/2020.lrec-1.6} {An annotated dataset of coreference in {E}nglish literature}.
\newblock In \emph{Proceedings of the Twelfth Language Resources and Evaluation Conference}, pages 44--54, Marseille, France. European Language Resources Association.

\bibitem[{Beltagy et~al.(2020)Beltagy, Peters, and Cohan}]{beltagy2020longformer}
Iz~Beltagy, Matthew~E Peters, and Arman Cohan. 2020.
\newblock Longformer: The long-document transformer.
\newblock \emph{arXiv preprint arXiv:2004.05150}.

\bibitem[{Botha et~al.(2020)Botha, Shan, and Gillick}]{botha-etal-2020-entity-100-languages}
Jan~A. Botha, Zifei Shan, and Daniel Gillick. 2020.
\newblock \href {https://doi.org/10.18653/v1/2020.emnlp-main.630} {{E}ntity {L}inking in 100 {L}anguages}.
\newblock In \emph{Proceedings of the 2020 Conference on Empirical Methods in Natural Language Processing (EMNLP)}, pages 7833--7845, Online. Association for Computational Linguistics.

\bibitem[{Chen et~al.(2021)Chen, Gudipati, Longpre, Ling, and Singh}]{chen-etal-2021-evaluating}
Anthony Chen, Pallavi Gudipati, Shayne Longpre, Xiao Ling, and Sameer Singh. 2021.
\newblock \href {https://doi.org/10.18653/v1/2021.acl-long.345} {Evaluating entity disambiguation and the role of popularity in retrieval-based {NLP}}.
\newblock In \emph{Proceedings of the 59th Annual Meeting of the Association for Computational Linguistics and the 11th International Joint Conference on Natural Language Processing (Volume 1: Long Papers)}, pages 4472--4485, Online. Association for Computational Linguistics.

\bibitem[{Chen et~al.(2017)Chen, Zhou, and Choi}]{chen-etal-2017-robust}
Henry~Y. Chen, Ethan Zhou, and Jinho~D. Choi. 2017.
\newblock \href {https://doi.org/10.18653/v1/K17-1023} {Robust coreference resolution and entity linking on dialogues: Character identification on {TV} show transcripts}.
\newblock In \emph{Proceedings of the 21st Conference on Computational Natural Language Learning ({C}o{NLL} 2017)}, pages 216--225, Vancouver, Canada. Association for Computational Linguistics.

\bibitem[{Chen et~al.(2018)Chen, Fan, Lu, Yuille, and Rong}]{preco-chen-etal-2018-preco}
Hong Chen, Zhenhua Fan, Hao Lu, Alan Yuille, and Shu Rong. 2018.
\newblock \href {https://doi.org/10.18653/v1/D18-1016} {{P}re{C}o: A large-scale dataset in preschool vocabulary for coreference resolution}.
\newblock In \emph{Proceedings of the 2018 Conference on Empirical Methods in Natural Language Processing}, pages 172--181, Brussels, Belgium. Association for Computational Linguistics.

\bibitem[{{Cologne University}(2024)}]{Cologne_Digital_Sanskrit_Dictionaries}
{Cologne University}. 2024.
\newblock \href {https://www.sanskrit-lexicon.uni-koeln.de} {Cologne digital sanskrit dictionaries, version 2.7.91}.
\newblock Accessed on January 30, 2024.

\bibitem[{Deleger et~al.(2012)Deleger, Li, Lingren, Kaiser, Molnar, Stoutenborough, Kouril, Marsolo, and Solti}]{Deleger2012}
Louise Deleger, Qi~Li, Todd Lingren, Megan Kaiser, Katalin Molnar, Laura Stoutenborough, Michal Kouril, Keith Marsolo, and Imre Solti. 2012.
\newblock \href {https://www.ncbi.nlm.nih.gov/pmc/articles/PMC3540456/} {Building gold standard corpora for medical natural language processing tasks}.
\newblock \emph{AMIA Annual Symposium Proceedings}, 2012:144--153.
\newblock PMID: 23304283; PMCID: PMC3540456.

\bibitem[{Durrett and Klein(2014)}]{durrett-klein-2014-joint}
Greg Durrett and Dan Klein. 2014.
\newblock \href {https://doi.org/10.1162/tacl_a_00197} {A joint model for entity analysis: Coreference, typing, and linking}.
\newblock \emph{Transactions of the Association for Computational Linguistics}, 2:477--490.

\bibitem[{Dwaipāyana and Duttā(1895)}]{krishna1895mahabharata}
Krishna Dwaipāyana and Manmatha~Nāth Duttā. 1895.
\newblock \emph{Mahābhārata}.
\newblock Elysium Press, Calcutta.

\bibitem[{F{\'e}vry et~al.(2020)F{\'e}vry, Baldini~Soares, FitzGerald, Choi, and Kwiatkowski}]{fevry-etal-2020-entities}
Thibault F{\'e}vry, Livio Baldini~Soares, Nicholas FitzGerald, Eunsol Choi, and Tom Kwiatkowski. 2020.
\newblock \href {https://doi.org/10.18653/v1/2020.emnlp-main.400} {Entities as experts: Sparse memory access with entity supervision}.
\newblock In \emph{Proceedings of the 2020 Conference on Empirical Methods in Natural Language Processing (EMNLP)}, pages 4937--4951, Online. Association for Computational Linguistics.

\bibitem[{Ghaddar and Langlais(2016)}]{ghaddar-langlais-2016-wikicoref}
Abbas Ghaddar and Phillippe Langlais. 2016.
\newblock \href {https://aclanthology.org/L16-1021} {{W}iki{C}oref: An {E}nglish coreference-annotated corpus of {W}ikipedia articles}.
\newblock In \emph{Proceedings of the Tenth International Conference on Language Resources and Evaluation ({LREC}'16)}, pages 136--142, Portoro{\v{z}}, Slovenia. European Language Resources Association (ELRA).

\bibitem[{Goyal and Huet(2016)}]{shr-Goyal_Huet_2016}
Pawan Goyal and Gerard Huet. 2016.
\newblock \href {https://doi.org/10.15398/jlm.v4i2.108} {Design and analysis of a lean interface for sanskrit corpus annotation}.
\newblock \emph{Journal of Language Modelling}, 4(2):145–182.

\bibitem[{Guo et~al.(2023)Guo, Hu, Zhang, Qiu, and Zhang}]{dualcache-guo-etal-2023-dual}
Qipeng Guo, Xiangkun Hu, Yue Zhang, Xipeng Qiu, and Zheng Zhang. 2023.
\newblock \href {https://doi.org/10.18653/v1/2023.acl-long.851} {Dual cache for long document neural coreference resolution}.
\newblock In \emph{Proceedings of the 61st Annual Meeting of the Association for Computational Linguistics (Volume 1: Long Papers)}, pages 15272--15285, Toronto, Canada. Association for Computational Linguistics.

\bibitem[{Han et~al.(2021)Han, Seo, Kang, Kim, Choi, Song, and Choi}]{fantasycoref-han-etal-2021-fantasycoref}
Sooyoun Han, Sumin Seo, Minji Kang, Jongin Kim, Nayoung Choi, Min Song, and Jinho~D. Choi. 2021.
\newblock \href {https://doi.org/10.18653/v1/2021.crac-1.3} {{F}antasy{C}oref: Coreference resolution on fantasy literature through omniscient writer{'}s point of view}.
\newblock In \emph{Proceedings of the Fourth Workshop on Computational Models of Reference, Anaphora and Coreference}, pages 24--35, Punta Cana, Dominican Republic. Association for Computational Linguistics.

\bibitem[{He et~al.(2013)He, Barbosa, and Kondrak}]{he-etal-2013-identification}
Hua He, Denilson Barbosa, and Grzegorz Kondrak. 2013.
\newblock \href {https://aclanthology.org/P13-1129/} {Identification of speakers in novels}.
\newblock In \emph{Proceedings of the 51st Annual Meeting of the Association for Computational Linguistics (Volume 1: Long Papers)}, pages 1312--1320, Sofia, Bulgaria. Association for Computational Linguistics.

\bibitem[{Hellwig and Nehrdich(2018)}]{hellwig-nehrdich-2018-sanskrit}
Oliver Hellwig and Sebastian Nehrdich. 2018.
\newblock \href {https://doi.org/10.18653/v1/D18-1295} {{S}anskrit word segmentation using character-level recurrent and convolutional neural networks}.
\newblock In \emph{Proceedings of the 2018 Conference on Empirical Methods in Natural Language Processing}, pages 2754--2763, Brussels, Belgium. Association for Computational Linguistics.

\bibitem[{Hoffart et~al.(2011)Hoffart, Yosef, Bordino, F{\"u}rstenau, Pinkal, Spaniol, Taneva, Thater, and Weikum}]{hoffart-etal-2011-robust}
Johannes Hoffart, Mohamed~Amir Yosef, Ilaria Bordino, Hagen F{\"u}rstenau, Manfred Pinkal, Marc Spaniol, Bilyana Taneva, Stefan Thater, and Gerhard Weikum. 2011.
\newblock \href {https://aclanthology.org/D11-1072/} {Robust disambiguation of named entities in text}.
\newblock In \emph{Proceedings of the 2011 Conference on Empirical Methods in Natural Language Processing}, pages 782--792, Edinburgh, Scotland, UK. Association for Computational Linguistics.

\bibitem[{Joshi et~al.(2019)Joshi, Levy, Zettlemoyer, and Weld}]{joshi-etal-2019-bert}
Mandar Joshi, Omer Levy, Luke Zettlemoyer, and Daniel Weld. 2019.
\newblock \href {https://doi.org/10.18653/v1/D19-1588} {{BERT} for coreference resolution: Baselines and analysis}.
\newblock In \emph{Proceedings of the 2019 Conference on Empirical Methods in Natural Language Processing and the 9th International Joint Conference on Natural Language Processing (EMNLP-IJCNLP)}, pages 5803--5808, Hong Kong, China. Association for Computational Linguistics.

\bibitem[{Jurafsky and Martin(2000)}]{jm3}
Daniel Jurafsky and James~H. Martin. 2000.
\newblock \emph{Speech and Language Processing: An Introduction to Natural Language Processing, Computational Linguistics, and Speech Recognition}, 1st edition.
\newblock Prentice Hall PTR, USA.

\bibitem[{Khanuja et~al.(2021)Khanuja, Bansal, Mehtani, Khosla, Dey, Gopalan, Margam, Aggarwal, Nagipogu, Dave, Gupta, Gali, Subramanian, and Talukdar}]{khanuja2021murilmultilingualrepresentationsindian}
Simran Khanuja, Diksha Bansal, Sarvesh Mehtani, Savya Khosla, Atreyee Dey, Balaji Gopalan, Dilip~Kumar Margam, Pooja Aggarwal, Rajiv~Teja Nagipogu, Shachi Dave, Shruti Gupta, Subhash Chandra~Bose Gali, Vish Subramanian, and Partha Talukdar. 2021.
\newblock \href {http://arxiv.org/abs/2103.10730} {Muril: Multilingual representations for indian languages}.

\bibitem[{Kim et~al.(2024)Kim, Lee, and Lee}]{kim2024koconovel}
Kyuhee Kim, Surin Lee, and Sangah Lee. 2024.
\newblock Koconovel: Annotated dataset of character coreference in korean novels.
\newblock \emph{arXiv preprint arXiv:2404.01140}.

\bibitem[{Krishna et~al.(2021)Krishna, Santra, Gupta, Satuluri, and Goyal}]{sanskrit-graphbased-10.1162/coli_a_00390}
Amrith Krishna, Bishal Santra, Ashim Gupta, Pavankumar Satuluri, and Pawan Goyal. 2021.
\newblock \href {https://doi.org/10.1162/coli_a_00390} {{A Graph-Based Framework for Structured Prediction Tasks in Sanskrit}}.
\newblock \emph{Computational Linguistics}, 46(4):785--845.

\bibitem[{Krug et~al.(2018)Krug, Puppe, Reger, Weimer, Macharowsky, Feldhaus, and Jannidis}]{krug2018droc}
Markus Krug, Frank Puppe, Isabella Reger, Lukas Weimer, Luisa Macharowsky, Stephan Feldhaus, and Fotis Jannidis. 2018.
\newblock \href {http://resolver.sub.uni-goettingen.de/purl/?dariah-2018-2} {Description of a corpus of character references in german novels - droc [deutsches roman corpus]}.
\newblock \emph{DARIAH-DE Working Papers}.

\bibitem[{Kummerfeld and Klein(2013)}]{kummerfeld-klein-2013-error}
Jonathan~K. Kummerfeld and Dan Klein. 2013.
\newblock \href {https://aclanthology.org/D13-1027} {Error-driven analysis of challenges in coreference resolution}.
\newblock In \emph{Proceedings of the 2013 Conference on Empirical Methods in Natural Language Processing}, pages 265--277, Seattle, Washington, USA. Association for Computational Linguistics.

\bibitem[{Lee et~al.(2017)Lee, He, Lewis, and Zettlemoyer}]{lee-etal-2017-end-to-end}
Kenton Lee, Luheng He, Mike Lewis, and Luke Zettlemoyer. 2017.
\newblock \href {https://doi.org/10.18653/v1/D17-1018} {End-to-end neural coreference resolution}.
\newblock In \emph{Proceedings of the 2017 Conference on Empirical Methods in Natural Language Processing}, pages 188--197, Copenhagen, Denmark. Association for Computational Linguistics.

\bibitem[{Likic(2008)}]{likic2008needleman}
Vladimir Likic. 2008.
\newblock The needleman-wunsch algorithm for sequence alignment.
\newblock \emph{Lecture given at the 7th Melbourne Bioinformatics Course, Bi021 Molecular Science and Biotechnology Institute, University of Melbourne}, pages 1--46.

\bibitem[{Limkonchotiwat et~al.(2023)Limkonchotiwat, Cheng, Christodoulopoulos, Saffari, and Lehmann}]{limkonchotiwat-etal-2023-mrefined}
Peerat Limkonchotiwat, Weiwei Cheng, Christos Christodoulopoulos, Amir Saffari, and Jens Lehmann. 2023.
\newblock \href {https://doi.org/10.18653/v1/2023.findings-emnlp.1007} {m{R}e{F}in{ED}: An efficient end-to-end multilingual entity linking system}.
\newblock In \emph{Findings of the Association for Computational Linguistics: EMNLP 2023}, pages 15080--15089, Singapore. Association for Computational Linguistics.

\bibitem[{Moosavi and Strube(2016)}]{moosavi-strube-2016-coreference-metric}
Nafise~Sadat Moosavi and Michael Strube. 2016.
\newblock \href {https://doi.org/10.18653/v1/P16-1060} {Which coreference evaluation metric do you trust? a proposal for a link-based entity aware metric}.
\newblock In \emph{Proceedings of the 54th Annual Meeting of the Association for Computational Linguistics (Volume 1: Long Papers)}, pages 632--642, Berlin, Germany. Association for Computational Linguistics.

\bibitem[{Moosavi and Strube(2017)}]{moosavi-strube-2017-lexical}
Nafise~Sadat Moosavi and Michael Strube. 2017.
\newblock \href {https://doi.org/10.18653/v1/P17-2003} {Lexical features in coreference resolution: To be used with caution}.
\newblock In \emph{Proceedings of the 55th Annual Meeting of the Association for Computational Linguistics (Volume 2: Short Papers)}, pages 14--19, Vancouver, Canada. Association for Computational Linguistics.

\bibitem[{Nedoluzhko et~al.(2022)Nedoluzhko, Nov{\'a}k, Popel, {\v{Z}}abokrtsk{\'y}, Zeldes, and Zeman}]{nedoluzhko-etal-2022-corefud}
Anna Nedoluzhko, Michal Nov{\'a}k, Martin Popel, Zden{\v{e}}k {\v{Z}}abokrtsk{\'y}, Amir Zeldes, and Daniel Zeman. 2022.
\newblock \href {https://aclanthology.org/2022.lrec-1.520} {{C}oref{UD} 1.0: Coreference meets {U}niversal {D}ependencies}.
\newblock In \emph{Proceedings of the Thirteenth Language Resources and Evaluation Conference}, pages 4859--4872, Marseille, France. European Language Resources Association.

\bibitem[{Nivre et~al.(2017)Nivre, Zeman, Ginter, and Tyers}]{nivre-etal-2017-universal}
Joakim Nivre, Daniel Zeman, Filip Ginter, and Francis Tyers. 2017.
\newblock \href {https://aclanthology.org/E17-5001/} {{U}niversal {D}ependencies}.
\newblock In \emph{Proceedings of the 15th Conference of the {E}uropean Chapter of the Association for Computational Linguistics: Tutorial Abstracts}, Valencia, Spain. Association for Computational Linguistics.

\bibitem[{Otmazgin et~al.(2023)Otmazgin, Cattan, and Goldberg}]{otmazgin-etal-2023-lingmess}
Shon Otmazgin, Arie Cattan, and Yoav Goldberg. 2023.
\newblock \href {https://doi.org/10.18653/v1/2023.eacl-main.202} {{L}ing{M}ess: Linguistically informed multi expert scorers for coreference resolution}.
\newblock In \emph{Proceedings of the 17th Conference of the European Chapter of the Association for Computational Linguistics}, pages 2752--2760, Dubrovnik, Croatia. Association for Computational Linguistics.

\bibitem[{Pagel and Reiter(2020)}]{pagel-reiter-2020-gerdracor}
Janis Pagel and Nils Reiter. 2020.
\newblock \href {https://aclanthology.org/2020.lrec-1.7} {{G}er{D}ra{C}or-coref: A coreference corpus for dramatic texts in {G}erman}.
\newblock In \emph{Proceedings of the Twelfth Language Resources and Evaluation Conference}, pages 55--64, Marseille, France. European Language Resources Association.

\bibitem[{Pradhan et~al.(2012)Pradhan, Moschitti, Xue, Uryupina, and Zhang}]{conll-pradhan-etal-2012-conll}
Sameer Pradhan, Alessandro Moschitti, Nianwen Xue, Olga Uryupina, and Yuchen Zhang. 2012.
\newblock \href {https://aclanthology.org/W12-4501} {{C}o{NLL}-2012 shared task: Modeling multilingual unrestricted coreference in {O}nto{N}otes}.
\newblock In \emph{Joint Conference on {EMNLP} and {C}o{NLL} - Shared Task}, pages 1--40, Jeju Island, Korea. Association for Computational Linguistics.

\bibitem[{Rao et~al.(2013)Rao, McNamee, and Dredze}]{Rao2013}
Delip Rao, Paul McNamee, and Mark Dredze. 2013.
\newblock \href {https://doi.org/10.1007/978-3-642-28569-1_5} {\emph{Entity Linking: Finding Extracted Entities in a Knowledge Base}}, pages 93--115. Springer Berlin Heidelberg, Berlin, Heidelberg.

\bibitem[{Roesiger et~al.(2018)Roesiger, Schulz, and Reiter}]{literary-text-roesiger-etal-2018-towards}
Ina Roesiger, Sarah Schulz, and Nils Reiter. 2018.
\newblock \href {https://aclanthology.org/W18-4515} {Towards coreference for literary text: Analyzing domain-specific phenomena}.
\newblock In \emph{Proceedings of the Second Joint {SIGHUM} Workshop on Computational Linguistics for Cultural Heritage, Social Sciences, Humanities and Literature}, pages 129--138, Santa Fe, New Mexico. Association for Computational Linguistics.

\bibitem[{S{\o}rensen(1904)}]{sorensen1904index}
S{\o}ren S{\o}rensen. 1904.
\newblock \emph{An Index to the Names in the Mahabharata: With Short Explanations and a Concordance to the Bombay and Calcutta Editions and P.C. Roy's Translation}, volume~1.
\newblock Williams \& Norgate, London.

\bibitem[{Stassen(1994)}]{Stassen_1994_copula}
Leon Stassen. 1994.
\newblock \href {https://doi.org/10.1017/S0332586500002961} {Typology versus mythology: The case of the zero-copula}.
\newblock \emph{Nordic Journal of Linguistics}, 17(2):105–126.

\bibitem[{Sujoy et~al.(2023)Sujoy, Krishna, and Goyal}]{sarkar-etal-2023-pre}
Sarkar Sujoy, Amrith Krishna, and Pawan Goyal. 2023.
\newblock \href {https://aclanthology.org/2023.wsc-csdh.4} {Pre-annotation based approach for development of a {S}anskrit named entity recognition dataset}.
\newblock In \emph{Proceedings of the Computational {S}anskrit {\&} Digital Humanities: Selected papers presented at the 18th World {S}anskrit Conference}, pages 59--70, Canberra, Australia (Online mode). Association for Computational Linguistics.

\bibitem[{Tjong Kim~Sang and De~Meulder(2003)}]{conll-ner-tjong-kim-sang-de-meulder-2003-introduction}
Erik~F. Tjong Kim~Sang and Fien De~Meulder. 2003.
\newblock \href {https://aclanthology.org/W03-0419} {Introduction to the {C}o{NLL}-2003 shared task: Language-independent named entity recognition}.
\newblock In \emph{Proceedings of the Seventh Conference on Natural Language Learning at {HLT}-{NAACL} 2003}, pages 142--147.

\bibitem[{Toshniwal et~al.(2020)Toshniwal, Wiseman, Ettinger, Livescu, and Gimpel}]{toshniwal-etal-2020-learning-to-ignore}
Shubham Toshniwal, Sam Wiseman, Allyson Ettinger, Karen Livescu, and Kevin Gimpel. 2020.
\newblock \href {https://doi.org/10.18653/v1/2020.emnlp-main.685} {Learning to {I}gnore: {L}ong {D}ocument {C}oreference with {B}ounded {M}emory {N}eural {N}etworks}.
\newblock In \emph{Proceedings of the 2020 Conference on Empirical Methods in Natural Language Processing (EMNLP)}, pages 8519--8526, Online. Association for Computational Linguistics.

\bibitem[{Tsai et~al.(2024)Tsai, Upadhyay, and Roth}]{Tsai2024}
Chen-Tse Tsai, Shyam Upadhyay, and Dan Roth. 2024.
\newblock \href {https://doi.org/10.1007/978-3-031-74901-8_1} {\emph{Introduction to Entity Discovery and Linking}}, pages 1--14. Springer Nature Switzerland, Cham.

\bibitem[{Uryupina et~al.(2020)Uryupina, Artstein, Bristot, Cavicchio, Delogu, Rodriguez, and Poesio}]{ARRAU_Uryupina_Artstein_Bristot_Cavicchio_Delogu_Rodriguez_Poesio_2020}
Olga Uryupina, Ron Artstein, Antonella Bristot, Federica Cavicchio, Francesca Delogu, Kepa~J. Rodriguez, and Massimo Poesio. 2020.
\newblock \href {https://doi.org/10.1017/S1351324919000056} {Annotating a broad range of anaphoric phenomena, in a variety of genres: The arrau corpus}.
\newblock \emph{Natural Language Engineering}, 26(1):95--128.

\bibitem[{{van Cranenburgh} and {van Noord}(2022)}]{openboek}
Andreas {van Cranenburgh} and Gertjan {van Noord}. 2022.
\newblock Openboek: A corpus of literary coreference and entities with an exploration of historical spelling normalization.
\newblock \emph{Computational Linguistics in the Netherlands Journal}, 12:235–251.

\bibitem[{{van Zundert} et~al.(2023){van Zundert}, {van Cranenburgh}, and Smeets}]{dutchcoref-van-Zundert}
Joris {van Zundert}, Andreas {van Cranenburgh}, and Roel Smeets. 2023.
\newblock Putting dutchcoref to the test: Character detection and gender dynamics in contemporary dutch novels.
\newblock In \emph{Proceedings of the Computational Humanities Research conference 2023}, pages 757--771. CEUR Workshop Proceedings (CEUR-WS.org).
\newblock Computational Humanities Research Conference ; Conference date: 06-12-2023 Through 08-12-2023.

\bibitem[{Wacks(2007)}]{wacks2007framing}
D.~Wacks. 2007.
\newblock \href {https://books.google.co.in/books?id=3xNGdR-pL50C} {\emph{Framing Iberia: Maq?m?t and Frametale Narratives in Medieval Spain}}.
\newblock The Medieval and Early Modern Iberian World. Brill.

\bibitem[{Yu et~al.(2020)Yu, Moosavi, Paun, and Poesio}]{yu-etal-2020-free-plural}
Juntao Yu, Nafise~Sadat Moosavi, Silviu Paun, and Massimo Poesio. 2020.
\newblock \href {https://doi.org/10.18653/v1/2020.coling-main.538} {Free the plural: Unrestricted split-antecedent anaphora resolution}.
\newblock In \emph{Proceedings of the 28th International Conference on Computational Linguistics}, pages 6113--6125, Barcelona, Spain (Online). International Committee on Computational Linguistics.

\bibitem[{Zhou and Choi(2018)}]{zhou-choi-2018-exist}
Ethan Zhou and Jinho~D. Choi. 2018.
\newblock \href {https://aclanthology.org/C18-1003/} {They exist! introducing plural mentions to coreference resolution and entity linking}.
\newblock In \emph{Proceedings of the 27th International Conference on Computational Linguistics}, pages 24--34, Santa Fe, New Mexico, USA. Association for Computational Linguistics.

\bibitem[{Özge Sevgili et~al.(2022)Özge Sevgili, Shelmanov, Arkhipov, Panchenko, and Biemann}]{ozge-Sevgili}
Özge Sevgili, Artem Shelmanov, Mikhail Arkhipov, Alexander Panchenko, and Chris Biemann. 2022.
\newblock \href {https://doi.org/10.3233/SW-222986} {Neural entity linking: A survey of models based on deep learning}.
\newblock \emph{Semantic Web}, 13(3):527--570.

\end{thebibliography}

\appendix

\section{Annotation Mapping Process}
\label{sec:appendix-2}
\begin{figure*}[htbp]
\centering
\includegraphics[width=\textwidth]{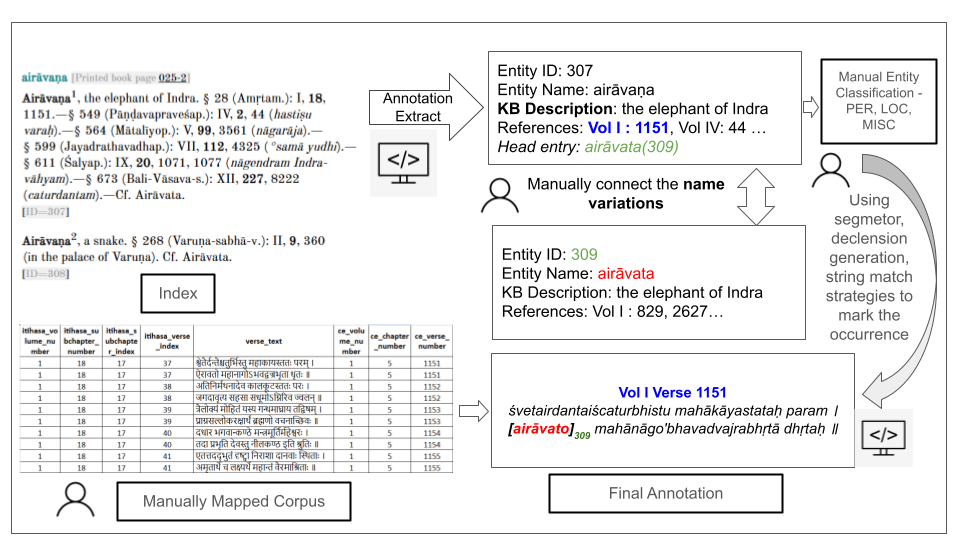}
\caption{The annotation pipeline for mapping index entries to the Itihasa Corpus. Entity variants are manually clustered, verse references are mapped to corpus verse IDs, and final occurrences are marked using a combination of string matching strategies.}
\label{fig:annotation-pipeline}
\end{figure*}

The process of creating our dataset, illustrated in Figure~\ref{fig:annotation-pipeline}, involved mapping the annotations provided by the "Index to the Names in the Mahabharata" to the "Itihasa Corpus". This process was divided into three main stages:

First, we extracted name variants and reference data from the index. As shown in the top-left of Figure~\ref{fig:annotation-pipeline}, each entry in the index includes multiple variant forms of a name, with associated verse references. We manually verified and connected these name variants to ensure accurate entity resolution (e.g., airāvana and airāvata).

Second, we aligned the verse numbers from the index—originally keyed to the Calcutta edition of the Mahabharata—with those used in the Itihasa Corpus. This required manually reading and mapping verse numbers to corresponding entries in the corpus (bottom-left of the figure).

Third, we marked the occurrences of each name within the corresponding verses. This was non-trivial because the index only lists verse numbers, not the exact token positions, and the textual data is unsegmented—meaning that names may appear compounded with other words in 39\% of cases.

To identify names within such tokens, we used the Sanskrit Heritage Reader (SHR), a lexicon-based shallow parser \cite{shr-Goyal_Huet_2016}, which could detect names in 85\% of cases by examining all valid segmentations. For 12\% of cases where SHR failed, we used a neural segmenter \cite{hellwig-nehrdich-2018-sanskrit}. In the remaining 3\%, where OCR errors or misspellings were present, we applied the Needleman-Wunsch approximate string matching algorithm \cite{likic2008needleman}, followed by manual correction. The final annotation, as seen at the bottom of Figure~\ref{fig:annotation-pipeline}, links each token-level name occurrence back to the correct Knowledge Bases entity ID.

\section{Entity Types and Examples}
\label{sec:appendix-entity-types}

Our annotation schema includes three coarse-grained entity types: \textbf{Person}, \textbf{Location}, and \textbf{Miscellaneous}. \textbf{Person} refers to named individuals or groups, human, personified, or divine, mentioned in the text, including relational mentions. \textbf{Location} includes named physical or conceptual places. \textbf{Miscellaneous} covers named objects, weapons, plants, any other names remaining in the index.

\begin{table}[!t]
\centering
\footnotesize
\begin{tabular}{|p{1.8cm}|p{5.0cm}|}
\hline
\textbf{Type} & \textbf{Example with Description} \\
\hline
\multirow{4}{*}{Person} & \textbf{Indra} – the chief of the devas, lord of rain \\
                        & \textbf{Aśvatthāman} – son of Droṇa and Kṛpī. \\
                        & \textbf{Madhusūdana} – alias of Kṛṣṇa \\
\hline
\multirow{3}{*}{Location} & \textbf{Kurukṣetra} – the country of the Kurus \\
                          & \textbf{Nīlaparvata} – a mountain \\
                          & \textbf{Brahmaloka} – the world of Brahman \\
\hline
\multirow{2}{*}{Misc} & \textbf{Sthuṇākarṇa} – name of a weapon \\
                      & \textbf{Mahāśaṅkha} – name of a tree \\
                      & \textbf{Kaumudī} -  the day of full moon in the month of Kaumuda \\
\hline
\end{tabular}
\caption{Examples of entity types.}
\label{tab:coarse-entities}
\end{table}

\section{Special Considerations}
\label{sec:appendix-1}
\textbf{Apposition and Copular Mentions:} Apposition occurs when two noun phrases refer to the same entity, with one providing additional information about the other. For example, in "kaunteyo dharmaputro yudhiṣṭhiraḥ" (Yudhishthira, the son of Kunti and Dharma), \textit{kaunteyo}, \textit{dharmaputro}, and \textit{yudhiṣṭhiraḥ} are coreferential \cite{nedoluzhko-etal-2022-corefud}. Copular mentions establish identity via a copula (e.g., "Yudhishthira is the son of Dharma"), but Sanskrit often omits it (zero-copula) due to its rich case system \cite{Stassen_1994_copula}. Following Preco \cite{preco-chen-etal-2018-preco} and KocoNovel \cite{kim2024koconovel}, we group appositive and copular mentions into the same cluster.

\textbf{Dual and Plural Mentions:}
Most coreference datasets assume anaphors have a single antecedent \cite{yu-etal-2020-free-plural}, with few exceptions like ARRAU \cite{ARRAU_Uryupina_Artstein_Bristot_Cavicchio_Delogu_Rodriguez_Poesio_2020}. Sanskrit also features a dual grammatical number, referring specifically to two entities. For example, \textit{mādrīputrau} and \textit{pāṇḍavau} refer to Nakula and Sahadeva. Following OntoNotes \cite{agarwal-etal-2022-entity-linking}, we mark dual and plural mentions as coreferential only with dual or plural antecedents.

\textbf{Nested Mentions: } 
Proper names are typically considered indivisible units, and any internal references within them are usually not annotated or identified \cite{kim2024koconovel}. Following this approach, we do not explicitly mark nested mentions as coreferential. For example, in \textit{dharmaputro} ("son of Dharma"), which refers to Yudhiṣṭhira, the nested entity \textit{dharma} ("the god of justice") is not separately annotated.

\textbf{Singletons:}
Singletons refer to entities with only one mention \cite{nedoluzhko-etal-2022-corefud}. Of the 5.5K entities in our dataset, 3.1K are singletons. As our dataset provides descriptions for all entities, and recent datasets such as LitBank \cite{litbank-bamman-etal-2020-annotated} and Preco \cite{preco-chen-etal-2018-preco} also include singletons for coreference tasks, we choose to keep the annotation for singletons.

\textbf{Unsegemeted Data: }
In Sanskrit, verses must adhere to one of the prescribed metrical patterns of Sanskrit prosody, which results in a relatively free word order, and words are often joined together to fit these metrical patterns \cite{sanskrit-graphbased-10.1162/coli_a_00390}. This leads to phonetic transformations (Sandhi) \cite{hellwig-nehrdich-2018-sanskrit}, merging words into continuous multi-word tokens. We keep the text unsegmented and mark entity boundaries at the character level rather than applying automatic segmentation \cite{hellwig-nehrdich-2018-sanskrit}. 39\% of mentions in our dataset consist of compounds or multi-word tokens.

\begin{flushleft}
\begin{enumerate}
    \item \textit{brahmaśiraḥ} + \textit{arjunena} \(\overset{\textit{aḥ + a = o'}}{\longrightarrow}\) \textit{brahmaśiro'rjunena} 
\end{enumerate}
\end{flushleft}

For example, in \textit{brahmaśiro'rjunena}, \textit{brahmaśiraḥ} ("Brahmashira weapon") and \textit{arjunena} ("by Arjuna") merge into a single span.

\section{Inter Annotator Agreement Study}
\label{sec:appendix-iaa}
To carry out the inter-annotator agreement (IAA) study, three groups independently annotated a set of 1,000 randomly selected verses using an online interface that supported both span marking and entity linking to Knowledge Base. Annotators were provided with verse numbers and access to the full corpus, enabling them to refer to broader narrative context when needed. The groups included two Sanskrit experts (both with master’s degrees, one with prior experience in \textit{Mahābhārata} studies) and a non-expert group with basic Sanskrit familiarity. All annotators had general cultural exposure to the epic. Agreement was measured by comparing each group’s annotations to ours using token-level Cohen’s $\kappa$ and F1 scores. For token-level $\kappa$, we computed agreement both over all tokens and over entity tokens only (i.e., tokens part of a mention by at least one annotator). F1 scores were calculated excluding non-entity labels, following guidelines by \citet{Deleger2012}.

While Cohen’s $\kappa$ remains a common IAA metric, it can be inflated in entity linking tasks due to token imbalance and sparse annotations. To address this, F1 scores which offers a more task-relevant view of agreement. Our annotations showed strong alignment with Expert 1 in both span detection and linking, with lower agreement observed for Expert 2 and the non-expert group—especially in the linking task. Notably, the F1 score difference between Expert 1 and Expert 2 for mention detection was modest (91 vs. 87), while the gap widened for entity linking (0.80 vs. 0.68), underscoring that disambiguation requires deeper domain understanding even among linguistically trained annotators.

\section{Implementation Details}
\label{sec:appendix-3}

We train our models using the Hugging Face library, initializing them with the Longformer-Large \cite{beltagy2020longformer}\footnote{\url{https://huggingface.co/allenai/longformer-large-4096}} and MuRIL \cite{khanuja2021murilmultilingualrepresentationsindian}\footnote{\url{https://huggingface.co/google/muril-base-cased}} pre-trained models. Our experiments involve three models: \textbf{LingMess}~\cite{otmazgin-etal-2023-lingmess}\footnote{\url{https://github.com/shon-otmazgin/lingmess-coref}}, \textbf{Dual Cache}~\cite{dualcache-guo-etal-2023-dual}\footnote{\url{https://github.com/QipengGuo/dual-cache-coref}}, and \textbf{mReFiNeD}~\cite{limkonchotiwat-etal-2023-mrefined}\footnote{\url{https://github.com/amazon-science/ReFinED}}.

\paragraph{LingMess.} We disable pronoun-related scorers (\texttt{PRON-PRON-C}, \texttt{PRON-PRON-NC}, \texttt{ENT-PRON}) as our dataset lacks pronoun annotations. The model is trained for 100 epochs on an NVIDIA L40 GPU, with hyperparameters tuned for validation F1-score. Training takes approximately 18 hours.

\paragraph{Dual Cache.} We configure the cache to prevent misses by setting the local cache (LRU) and global cache (LFU) sizes to 1000. The model is also trained for 100 epochs on an NVIDIA L40 GPU, and training requires around 34 hours.

\paragraph{mReFiNeD.} We train mReFiNeD in a multi-task setting for mention detection, entity typing, disambiguation, and linking. We use coarse-grained tags (\texttt{PER}, \texttt{LOC}, \texttt{MISC}) and retain 30 candidates per mention, which include the gold entity, the top-ranked candidate, and random negatives. Candidate ranking uses the estimated probability \(\hat{p}(e_j | m_i)\), with global priors estimated from the training corpus. Both mention and description encoders use MuRIL, a multilingual model for Indian languages. Training is done for 40 epochs on an NVIDIA A40 GPU and completes in approximately 8 hours.

We explore batch sizes of 8, 16, and 32 during hyperparameter search, while keeping other parameters aligned with the original model implementations.

\section{Quantitative Error Analysis}
\label{sec:appendix-error}

Table~\ref{tab:error} categorizes model-specific errors using the Berkeley Coreference Analyzer framework~\cite{kummerfeld-klein-2013-error}, adapted to our single-token mention setup. The following error types were considered: \textit{Conflated Entity}, where distinct gold entities are incorrectly merged; \textit{Divided Entity}, where a single gold entity is erroneously split into multiple predicted clusters; \textit{Missing Entity / Mention}, where the system fails to identify a gold entity or mention; and \textit{Extra Entity / Mention}, where the model predicts an entity or mention that does not exist in the gold annotations.

\textbf{Conflated Entity} errors (e.g., 10.4\% for Lingmess) occur when the model merges mentions of different entities. This aligns with the qualitative error noted where \textit{Bhūri} and \textit{Duryodhana} are both grouped under the common term \textit{kaurava}, leading to incorrect entity merging due to insufficient disambiguation. These errors are highest in Lingmess and lowest in mReFiNeD, as the latter was provided with a possible alias list based on prior probability.

\textbf{Divided Entity} errors (e.g., 33.2\% for Dual-Cache Global) reflect over-splitting of a single entity into multiple clusters. This supports our qualitative observation regarding \textit{Draupadī}, where lexical variants like \textit{Yājñasenī}, \textit{Kṛṣṇāṃ}, and \textit{Pāñcālyā} were not clustered together. These errors are highest in Dual-Cache Global, as the model struggled to connect all mentions of entities across the full test set, and lowest in mReFiNeD due to its use of a prior-based alias list.

\textbf{Missing and Extra Mentions/Entities} highlight the difficulty models face in detecting all valid references. For instance, \textbf{Extra Entity} errors peak at 37.7\% for Dual-Cache Global due to divided entities and the model's failure to align all entities, while \textbf{Missing Entity} errors reach 32.7\% for mReFiNeD due to poor mention detection in end-to-end training.

\section{Additional Dataset Statistics}
The dataset exhibits an average entity density of 0.11 (i.e., roughly one entity mention every ten tokens). Entity overlap across sections is limited: the average Jaccard similarity across chapters is 0.127, indicating that only about 13\% of entities, which are major entities, are shared between any two chapters. At the subchapter level, the average Jaccard 0.064.

\begin{figure}[t]
    \centering
    \includegraphics[width=\linewidth]{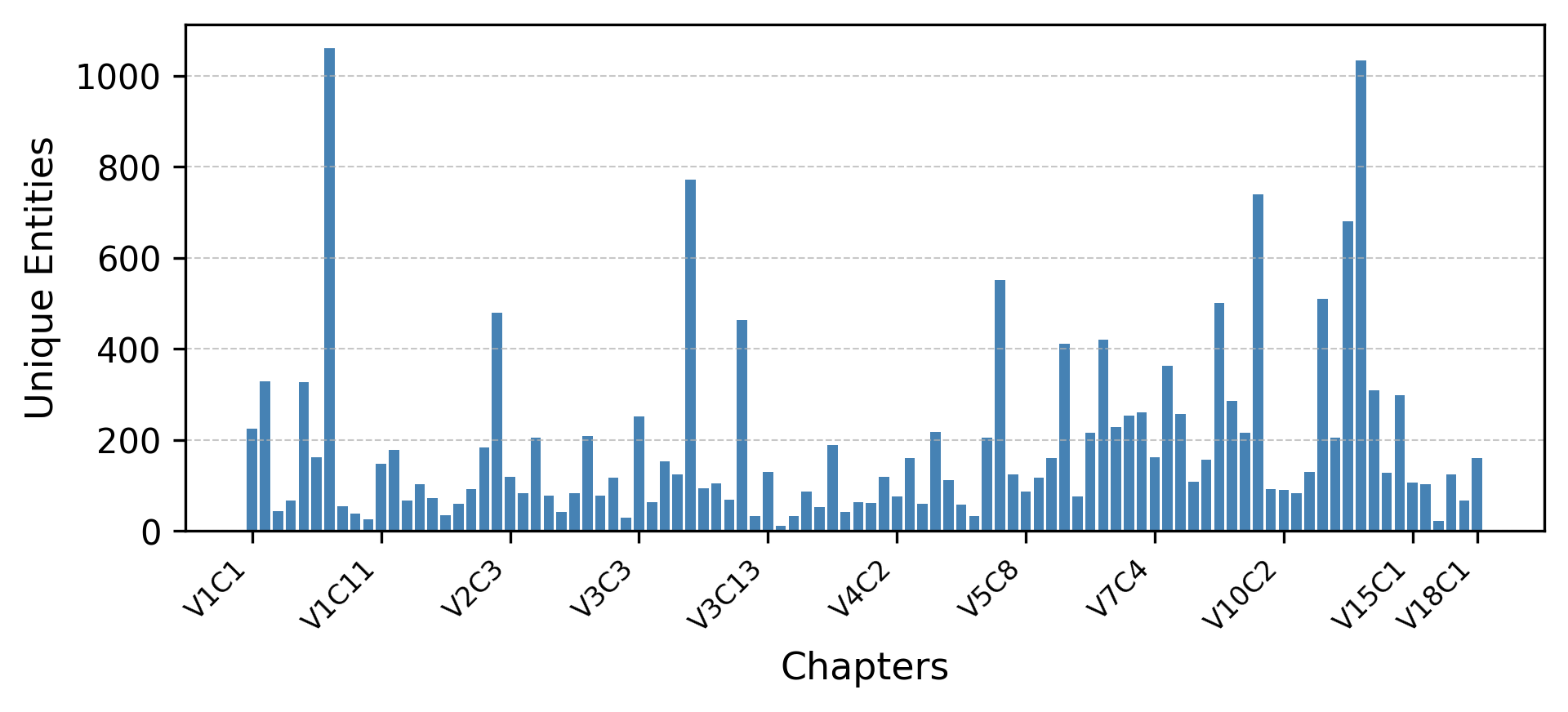}
    \caption{Distribution of unique entities per chapter in the dataset.}
    \label{fig:unique-entities}
\end{figure}

At the chapter level, the average chain length is 5.97, while at the subchapter level it is 2.48.  However, focusing on the major entities, their average chain length rises to 31.35 at the chapter level and 4.04 at the subchapter level. Thus, while many entities are ephemeral, a handful of central characters maintain long and recurrent chains that dominate the discourse.
Figure \ref{fig:unique-entities} show the distribution of unique entities across chapters.

\end{document}